\documentclass[11pt,letterpaper]{article}
\usepackage{naaclhlt2016}
\usepackage{times}
\usepackage{url}
\usepackage{latexsym}
\usepackage{multirow}
\usepackage{amssymb}
\usepackage{amsthm}
\usepackage{color}
\usepackage{latexsym}
\usepackage{amsmath}
\usepackage{verbatim}
\usepackage{tikz-qtree}
\usepackage{tikz}
\usetikzlibrary{shapes.geometric}
\usetikzlibrary{shapes.arrows}
\usetikzlibrary{patterns}
\usetikzlibrary{chains}
\usetikzlibrary{calc}

\naaclfinalcopy
\def\naaclpaperid{521}

\newcommand{\ignore}[1]{}

\newcommand{\miguelcomment}[1]{\ignore{\textcolor{red}{\textbf{[#1 --\textsc{MB}]}}}}
\newcommand{\guillaumecomment}[1]{\ignore{\textcolor{orange}{\textbf{[#1 --\textsc{GL}]}}}}

\newsavebox{\one}
\newsavebox{\two}
\newsavebox{\three}
\newsavebox{\four}
\newsavebox{\five}

\DeclareMathOperator{\logadd}{logadd}
\DeclareMathOperator*{\argmax}{\arg\!\max}

\title{Neural Architectures for Named Entity Recognition}
\author{Guillaume Lample$^{\spadesuit}$ ~ Miguel Ballesteros$^{\clubsuit\spadesuit}$ \\ ~ \textbf{Sandeep Subramanian$^{\spadesuit}$ ~ Kazuya Kawakami$^\spadesuit$ ~ Chris Dyer$^{\spadesuit}$}\\
  $^\spadesuit$Carnegie Mellon University ~~ $^\clubsuit$NLP Group, Pompeu Fabra University \\
 { \tt $\{$glample,sandeeps,kkawakam,cdyer\}@cs.cmu.edu, } \\ { \tt  miguel.ballesteros@upf.edu}
}
\date{}

\begin{document}

\maketitle

\begin{abstract}
State-of-the-art named entity recognition systems rely heavily on hand-crafted features and domain-specific knowledge in order to learn effectively from the small, supervised training corpora that are available. In this paper, we introduce two new neural architectures---one based on bidirectional LSTMs and conditional random fields, and the other that constructs and labels segments using a transition-based approach inspired by shift-reduce parsers. Our models rely on two sources of information about words: character-based word representations learned from the supervised corpus and unsupervised word representations learned from unannotated corpora. Our models obtain state-of-the-art performance in NER in four languages without resorting to any language-specific knowledge or resources such as gazetteers. \footnote{The code of the  LSTM-CRF and Stack-LSTM NER systems are available at \url{https://github.com/glample/tagger} and \url{https://github.com/clab/stack-lstm-ner}
}
\end{abstract}

\section{Introduction}

Named entity recognition (NER) is a challenging learning problem. One the one hand, in most languages and domains, there is only a very small amount of supervised training data available. On the other, there are few constraints on the kinds of words that can be names, so generalizing from this small sample of data is difficult. As a result, carefully constructed orthographic features and language-specific knowledge resources, such as gazetteers, are widely used for solving this task. Unfortunately, language-specific resources and features are costly to develop in new languages and new domains, making NER a challenge to adapt. Unsupervised learning from unannotated corpora offers an alternative strategy for obtaining better generalization from small amounts of supervision. However, even systems that have relied extensively on unsupervised features~\cite[\emph{inter alia}]{collobert2011natural,turian:2010,lin2009phrase,ando:2005} have used these to augment, rather than replace, hand-engineered features (e.g., knowledge about capitalization patterns and character classes in a particular language) and specialized knowledge resources (e.g., gazetteers).

In this paper, we present neural architectures for NER that use no language-specific resources or features beyond a small amount of supervised training data and unlabeled corpora. Our models are designed to capture two intuitions. First, since names often consist of multiple tokens, reasoning jointly over tagging decisions for each token is important. We compare two models here, (i)~a bidirectional LSTM with a sequential conditional random layer above it~(LSTM-CRF; \S\ref{lstmcrf}), and (ii)~a new model that constructs and labels chunks of input sentences using an algorithm inspired by transition-based parsing with states represented by stack LSTMs~(S-LSTM; \S\ref{stacklstm}). Second, token-level evidence for ``being a name'' includes both orthographic evidence (what does the word being tagged as a name look like?) and distributional evidence (where does the word being tagged tend to occur in a corpus?). To capture orthographic sensitivity, we use character-based word representation model~\cite{ling:2015} to capture distributional sensitivity, we combine these representations with distributional representations~\cite{mikolov2013distributed}. Our word representations combine both of these, and dropout training is used to encourage the model to learn to trust both sources of evidence~(\S\ref{sec:words}).


Experiments in English, Dutch, German, and Spanish show that we are able to obtain state-of-the-art NER performance with the LSTM-CRF model in Dutch, German, and Spanish, and very near the state-of-the-art in English without any hand-engineered features or gazetteers~(\S\ref{sec:experiments}). The transition-based algorithm likewise surpasses the best previously published results in several languages, although it performs less well than the LSTM-CRF model.
\guillaumecomment{actually we dont have sota on dutch if we include this recent byte model that uses external NER data :/}

\section{LSTM-CRF Model}
\label{lstmcrf}

We provide a brief description of LSTMs and CRFs, and present a hybrid tagging architecture. This architecture is similar to the ones presented by \newcite{collobert2011natural} and \newcite{huang:2015}.

\subsection{LSTM}
\label{sec:lstm}

Recurrent neural networks (RNNs) are a family of neural networks that operate on sequential data. They take as input a sequence of vectors $(\mathbf{x}_1, \mathbf{x}_2, \ldots, \mathbf{x}_n)$ and return another sequence $(\mathbf{h}_1, \mathbf{h}_2, \ldots, \mathbf{h}_n)$ that represents some information about the sequence at every step in the input. Although RNNs can, in theory, learn long dependencies, in practice they fail to do so and tend to be biased towards their most recent inputs in the sequence \cite{bengio1994learning}. Long Short-term Memory Networks (LSTMs) have been designed to combat this issue by incorporating a memory-cell and have been shown to capture long-range dependencies. They do so using several gates that control the proportion of the input to give to the memory cell, and the proportion from the previous state to forget~\cite{hochreiter:1997}.
We use the following implementation:
\\
\begin{align*}
\mathbf{i}_{t} &= \sigma(\mathbf{W}_{xi}\mathbf{x}_{t} + \mathbf{W}_{hi}\mathbf{h}_{t-1} + \mathbf{W}_{ci}\mathbf{c}_{t-1} + \mathbf{b}_{i})\\
\mathbf{c}_{t} &= (1 - \mathbf{i}_{t})\odot\mathbf{c}_{t-1} +\\
&\qquad \mathbf{i}_{t}\odot \tanh(\mathbf{W}_{xc}\mathbf{x}_{t} + \mathbf{W}_{hc}\mathbf{h}_{t-1} + \mathbf{b}_{c})\\
\mathbf{o}_{t} &= \sigma(\mathbf{W}_{xo}\mathbf{x}_{t} + \mathbf{W}_{ho}\mathbf{h}_{t-1} + \mathbf{W}_{co}\mathbf{c}_{t} + \mathbf{b}_{o})\\
\mathbf{h}_{t} &= \mathbf{o}_{t}\odot\tanh(\mathbf{c}_{t}),
\end{align*}
where $\sigma$ is the element-wise sigmoid function, and $\odot$ is the element-wise product.

For a given sentence $(\mathbf{x}_1, \mathbf{x}_2, \ldots, \mathbf{x}_n)$ containing $n$ words, each represented as a $d$-dimensional vector, an LSTM computes a representation $\overrightarrow{\mathbf{h}_t}$ of the left context of the sentence at every word $t$. Naturally, generating a representation of the right context $\overleftarrow{\mathbf{h}_t}$ as well should add useful information. This can be achieved using a second LSTM that reads the same sequence in reverse. We will refer to the former as the forward LSTM and the latter as the backward LSTM. These are two distinct networks with different parameters. This forward and backward LSTM pair is referred to as a bidirectional LSTM~\cite{graves:2005}.

The representation of a word using this model is obtained by concatenating its left and right context representations, $\mathbf{h}_{t} = [\overrightarrow{\mathbf{h}_{t}} ; \overleftarrow{\mathbf{h}_{t}}]$. These representations effectively include a representation of a word in context, which is useful for numerous tagging applications.

\subsection{CRF Tagging Models}
\label{sec:crf}

A very simple---but surprisingly effective---tagging model is to use the $\mathbf{h}_t$'s as features to make independent tagging decisions for each output $y_t$~\cite{ling:2015}. Despite this model's success in simple problems like POS tagging, its independent classification decisions are limiting when there are strong dependencies across output labels. NER is one such task, since the ``grammar'' that characterizes interpretable sequences of tags imposes several hard constraints (e.g., I-PER cannot follow B-LOC; see \S\ref{IOBES} for details) that would be impossible to model with independence assumptions.

Therefore, instead of modeling tagging decisions independently, we model them jointly using a conditional random field~\cite{lafferty2001conditional}. For an input sentence
$$\mathbf{X} = (\mathbf{x}_1, \mathbf{x}_2, \ldots, \mathbf{x}_n),$$
we consider $\mathbf{P}$ to be the matrix of scores output by the bidirectional LSTM network. $\mathbf{P}$ is of size $n~\times~k$, where $k$ is the number of distinct tags, and $P_{i, j}$ corresponds to the score of the $j^{th}$ tag of the $i^{th}$ word in a sentence. For a sequence of predictions
$$\mathbf{y} = (y_1, y_2, \ldots, y_n),$$
we define its score to be
$$s(\mathbf{X}, \mathbf{y})=\sum_{i=0}^{n} A_{y_i, y_{i+1}} + \sum_{i=1}^{n} P_{i, y_i}$$
where $\mathbf{A}$ is a matrix of transition scores such that $A_{i, j}$ represents the score of a transition from the tag $i$ to tag $j$. $y_0$ and $y_n$ are the \textit{start} and \textit{end} tags of a sentence, that we add to the set of possible tags. $\mathbf{A}$ is therefore a square matrix of size $k+2$.
\\
\\
A softmax over all possible tag sequences yields a probability for the sequence $\mathbf{y}$:
$$p(\mathbf{y} | \mathbf{X}) = \frac{
	e^{s(\mathbf{X}, \mathbf{y})}
}{
	\sum_{\mathbf{\widetilde{y}} \in \mathbf{Y_X}} e^{s(\mathbf{X}, \mathbf{\widetilde{y}})}
}.$$
During training, we maximize the log-probability of the correct tag sequence:
\\
\begin{align}
\log(p(\mathbf{y} | \mathbf{X})) &= s(\mathbf{X}, \mathbf{y}) - \log \left( \sum_{\mathbf{\widetilde{y}} \in \mathbf{Y_X}} e^{s(\mathbf{X}, \mathbf{\widetilde{y}})} \right) \nonumber \\
&= s(\mathbf{X}, \mathbf{y}) - \underset{{\mathbf{\widetilde{y}} \in \mathbf{Y_X}}}{\logadd}\ s(\mathbf{X}, \mathbf{\widetilde{y}}), \label{eq:crf}
\end{align}
where $\mathbf{Y_X}$ represents all possible tag sequences (even those that do not verify the IOB format) for a sentence $\mathbf{X}$. From the formulation above, it is evident that we encourage our network to produce a valid sequence of output labels. While decoding, we predict the output sequence that obtains the maximum score given by:
\begin{align}
\mathbf{y}^* = \argmax_{\mathbf{\widetilde{y}} \in \mathbf{Y_X}}{s(\mathbf{X}, \mathbf{\widetilde{y}})}. \label{eq:vit}
\end{align}

Since we are only modeling bigram interactions between outputs, both the summation in Eq.~\ref{eq:crf} and the maximum a posteriori sequence $\mathbf{y}^*$ in Eq.~\ref{eq:vit} can be computed using dynamic programming.

\subsection{Parameterization and Training}
\begin{figure}
  \centering
    \includegraphics[scale=0.64]{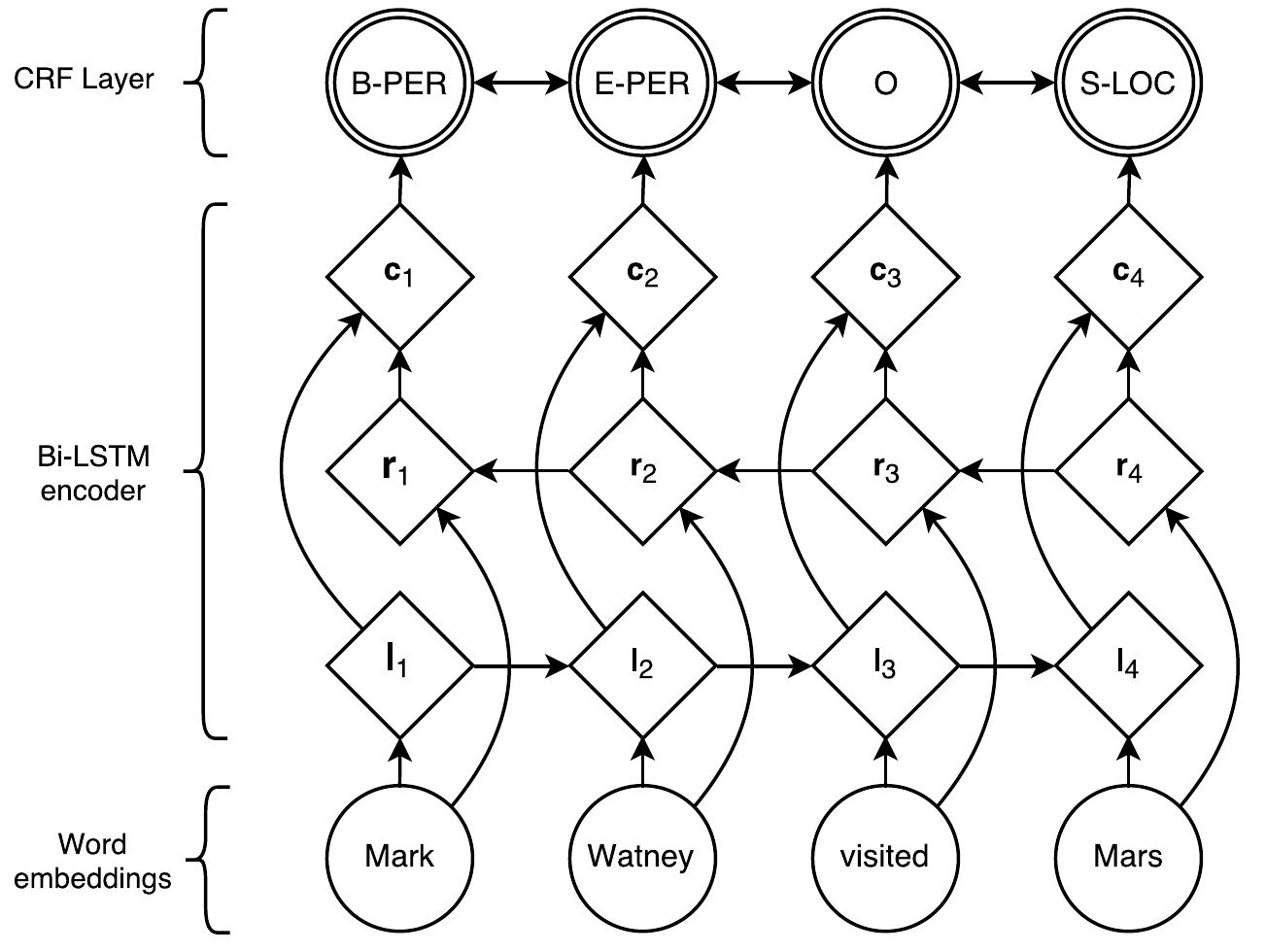}
  \caption{Main architecture of the network. Word embeddings are given to a bidirectional LSTM. $\mathbf{l}_i$ represents the word $i$ and its left context, $\mathbf{r}_i$ represents the word $i$ and its right context. Concatenating these two vectors yields a representation of the word $i$ in its context, $\mathbf{c}_i$.}
  \label{fig:bilstm-crf}
\end{figure}

The scores associated with each tagging decision for each token (i.e., the $P_{i,y}$'s) are defined to be the dot product between the embedding of a word-in-context computed with a bidirectional LSTM---exactly the same as the POS tagging model of \newcite{ling:2015} and these are combined with bigram compatibility scores (i.e., the $A_{y,y'}$'s). This architecture is shown in figure \ref{fig:bilstm-crf}. Circles represent observed variables, diamonds are deterministic functions of their parents, and double circles are random variables.

The parameters of this model are thus the matrix of bigram compatibility scores $\mathbf{A}$, and the parameters that give rise to the matrix $\mathbf{P}$, namely the parameters of the bidirectional LSTM, the linear feature weights, and the word embeddings. As in part~\ref{sec:crf}, let $\mathbf{x}_i$ denote the sequence of word embeddings for every word in a sentence, and $y_i$ be their associated tags. We return to a discussion of how the embeddings $\mathbf{x}_i$ are modeled in Section~\ref{sec:words}. The sequence of word embeddings is given as input to a bidirectional LSTM, which returns a representation of the left and right context for each word as explained in ~\ref{sec:lstm}.

These representations are concatenated ($\mathbf{c}_i$) and linearly projected onto a layer whose size is equal to the number of distinct tags. Instead of using the softmax output from this layer, we use a CRF as previously described to take into account neighboring tags, yielding the final predictions for every word $y_i$. Additionally, we observed that adding a hidden layer between $\mathbf{c}_i$ and the CRF layer marginally  improved our results. All results reported with this model incorporate this extra-layer.  The parameters are trained to maximize Eq.~\ref{eq:crf} of observed sequences of NER tags in an annotated corpus, given the observed words.

\subsection{Tagging Schemes}
\label{IOBES}
The task of named entity recognition is to assign a named entity label to every word in a sentence. A single named entity could span several tokens within a sentence. Sentences are usually represented in the IOB format (Inside, Outside, Beginning) where every token is labeled as B-\textit{label} if the token is the beginning of a named entity, I-\textit{label} if it is inside a named entity but not the first token within the named entity, or O otherwise. However, we decided to use the IOBES tagging scheme, a variant of IOB commonly used for named entity recognition, which encodes information about singleton entities (S) and explicitly marks the end of named entities (E). Using this scheme, tagging a word as I-\textit{label} with high-confidence narrows down the choices for the subsequent word to I-\textit{label} or E-\textit{label}, however, the IOB scheme is only capable of determining that the subsequent word cannot be the interior of another label. \newcite{ratinov2009design} and \newcite{dai2015enhancing} showed that using a more expressive tagging scheme like IOBES improves model performance marginally. However, we did not observe a significant improvement over the IOB tagging scheme. 

\section{Transition-Based Chunking Model}
\label{stacklstm}

As an alternative to the LSTM-CRF discussed in the previous section, we explore a new architecture that chunks and labels a sequence of inputs using an algorithm similar to transition-based dependency parsing. This model directly constructs representations of the multi-token names (e.g., the name \emph{Mark Watney} is composed into a single representation).

This model relies on a stack data structure to incrementally construct chunks of the input. To obtain representations of this stack used for predicting subsequent actions, we use the Stack-LSTM presented by \newcite{dyer:2015}, in which the LSTM is augmented with a ``stack pointer.'' While sequential LSTMs model sequences from left to right, stack LSTMs permit embedding of a stack of objects that are both added to (using a push operation) and removed from (using a pop operation). This allows the Stack-LSTM to work like a stack that maintains a ``summary embedding'' of its contents. We refer to this model as Stack-LSTM or S-LSTM model for simplicity. 

Finally, we refer interested readers to the original paper \cite{dyer:2015} for details about the Stack-LSTM model since in this paper we merely use the same architecture through a new transition-based algorithm presented in the following Section.

\subsection{Chunking Algorithm}

We designed a transition inventory which is given in Figure~\ref{fig:parser} that is inspired by transition-based parsers, in particular the arc-standard parser of \newcite{nivre2004}. In this algorithm, we make use of two stacks (designated \emph{output} and \emph{stack} representing, respectively, completed chunks and scratch space) and a \emph{buffer} that contains the words that have yet to be processed. The transition inventory contains the following transitions: The \textsc{shift} transition moves a word from the buffer to the stack, the \textsc{out} transition moves a word from the buffer directly into the output stack while the \textsc{reduce}$(y)$ transition pops all items from the top of the stack creating a ``chunk,'' labels this with label $y$, and pushes a representation of this chunk onto the output stack. The algorithm completes when the stack and buffer are both empty. The algorithm is depicted in Figure \ref{fig:parser}, which shows the sequence of operations required to process the sentence \emph{Mark Watney visited Mars}.

The model is parameterized by defining a probability distribution over actions at each time step, given the current contents of the stack, buffer, and output, as well as the history of actions taken. Following \newcite{dyer:2015}, we use stack LSTMs to compute a fixed dimensional embedding of each of these, and take a concatenation of these to obtain the full algorithm state. This representation is used to define a distribution over the possible actions that can be taken at each time step. The model is trained to maximize the conditional probability of sequences of reference actions (extracted from a labeled training corpus) given the input sentences. To label a new input sequence at test time, the maximum probability action is chosen greedily until the algorithm reaches a termination state. Although this is not guaranteed to find a global optimum, it is effective in practice. Since each token is either moved directly to the output (1 action) or first to the stack and then the output (2 actions), the total number of actions for a sequence of length $n$ is maximally $2n$.

\begin{figure*}
\begin{small}
\centering
\begin{tabular}{lll|l|lll|c}
\textbf{Out}$_t$ & \textbf{Stack}$_t$ & \textbf{Buffer}$_t$ & \textbf{Action} & \textbf{Out}$_{t+1}$ & \textbf{Stack}$_{t+1}$ & \textbf{Buffer}$_{t+1}$ & \textbf{Segments} \\
\hline
$O$ & $S$ & $(\mathbf{u},u),B$ & \textsc{shift} & $O$ & $(\mathbf{u},u),S$ & $B$ & --- \\ 
$O$ & $(\mathbf{u},u),\ldots,(\mathbf{v},v),S$ & $B$  &$\textsc{reduce}(y)$ & $g(\mathbf{u},\ldots,\mathbf{v},\mathbf{r}_y),O$ & $S$ & $B$ & $(u\ldots v,y)$ \\
$O$ & $S$ & $(\mathbf{u},u),B$ & \textsc{out} & $g(\mathbf{u},\mathbf{r}_{\varnothing}),O$ & $S$ & $B$ & ---
\end{tabular}
\end{small}
\caption{Transitions of the Stack-LSTM model indicating the action applied and the resulting state. \guillaumecomment{this first sentence is maybe a little bit confusing} Bold symbols indicate (learned) embeddings of words and relations, script symbols indicate the corresponding words and relations.}
\label{fig:parser}
\end{figure*}

\ignore{
\begin{figure}
  \centering
  \includegraphics[scale=0.44]{stack_lstm}
  \caption{Intermediate state computation for the sentence \emph{Mark Watney visited Mars} with the Stack-LSTM model.}
  \label{figurekazuya}
\end{figure}
}

\begin{figure*}[t]
  \begin{center}
    \centering
    \begin{scriptsize}
      \begin{tabular}{lllll}\textbf{Transition}&\textbf{Output}&\textbf{Stack}&\textbf{Buffer}&\textbf{Segment}\\
      \hline
        &[]&[]&[Mark, Watney, visited, Mars]&\\
        \textsc{Shift}&[]&[Mark]&[Watney, visited, Mars]&\\
        \textsc{Shift}&[]&[Mark, Watney]&[visited, Mars]&\\
        \textsc{REDUCE(PER)}&[(Mark Watney)-PER]&[]&[visited, Mars]& (Mark Watney)-PER\\
        \textsc{OUT}&[(Mark Watney)-PER, visited]&[]&[Mars]&\\
        \textsc{SHIFT}&[(Mark Watney)-PER, visited]&[Mars]&[]&\\
        \textsc{REDUCE(LOC)}&[(Mark Watney)-PER, visited, (Mars)-LOC]&[]&[]& (Mars)-LOC\\
      \end{tabular}
    \end{scriptsize}
    \caption{Transition sequence for \emph{Mark Watney visited Mars} with the Stack-LSTM model.}
    \label{parsingexample}
  \end{center}
\end{figure*}

It is worth noting that the nature of this algorithm model makes it agnostic to the tagging scheme used since it directly predicts labeled chunks.

\subsection{Representing Labeled Chunks}

When the $\textsc{reduce}(y)$ operation is executed, the algorithm shifts a sequence of tokens (together with their vector embeddings) from the stack to the output buffer as a single completed chunk. To compute an embedding of this sequence, we run a bidirectional LSTM over the embeddings of its constituent tokens together with a token representing the type of the chunk being identified (i.e., $y$). This function is given as $g(\mathbf{u}, \ldots, \mathbf{v},\mathbf{r}_y)$, where $\mathbf{r}_y$ is a learned embedding of a label type. Thus, the output buffer contains a single vector representation for each labeled chunk that is generated, regardless of its length.

\section{Input Word Embeddings}\label{sec:words}
The input layers to both of our models are vector representations of individual words. Learning independent representations for word types from the limited NER training data is a difficult problem: there are simply too many parameters to reliably estimate. Since many languages have orthographic or morphological evidence that something is a name (or not a name), we want representations that are sensitive to the spelling of words. We therefore use a model that constructs representations of words from representations of the characters they are composed of~(\ref{sec:character-model}). Our second intuition is that names, which may individually be quite varied, appear in regular contexts in large corpora. Therefore we use embeddings learned from a large corpus that are sensitive to word order~(\ref{sec:pretrained}). Finally, to prevent the models from depending on one representation or the other too strongly, we use dropout training and find this is crucial for good generalization performance~(\ref{sec:dropout}).

\subsection{Character-based models of words}
\label{sec:character-model}

\begin{figure}
  \centering
    \includegraphics[scale=0.68]{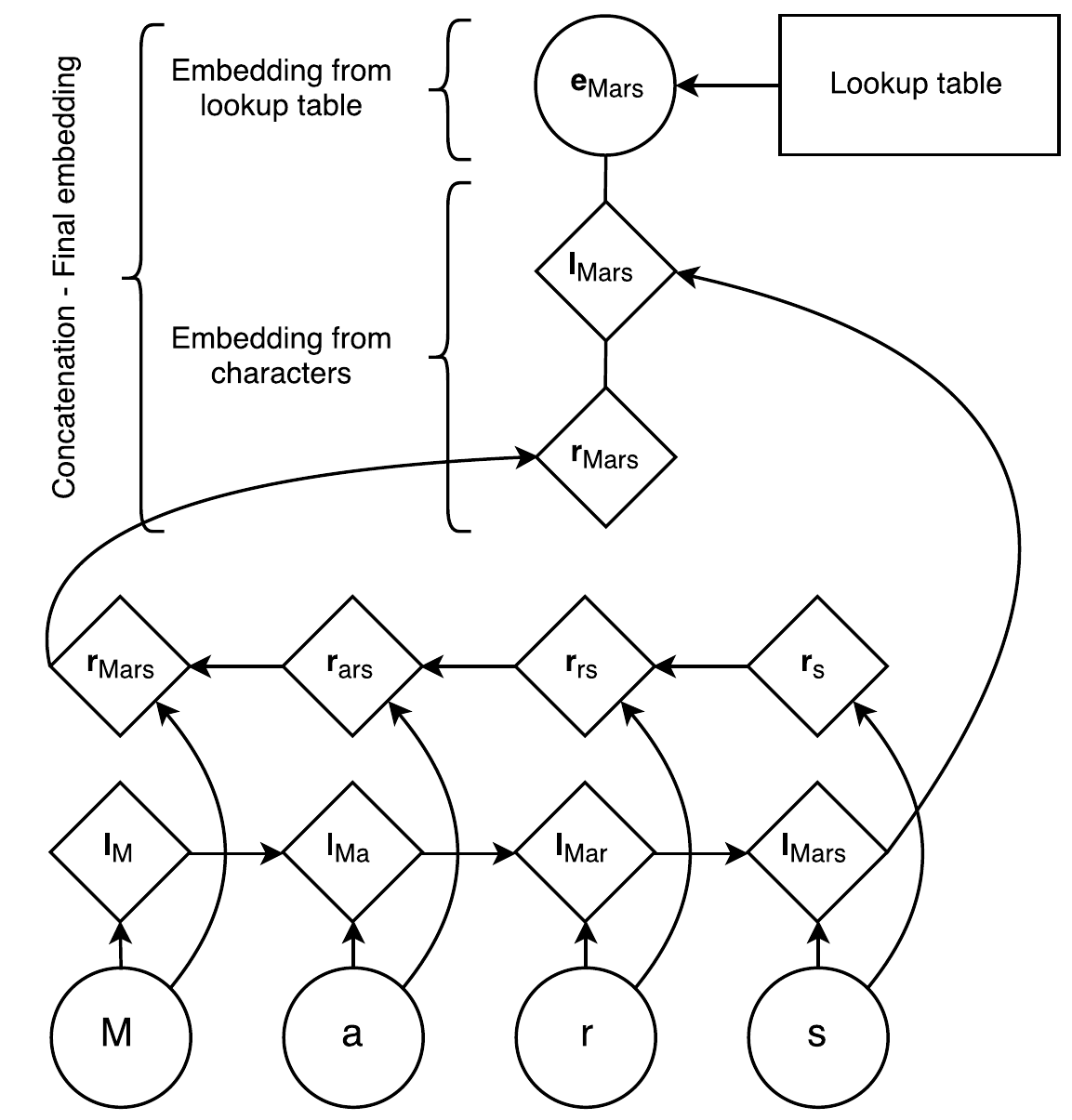}
  \caption{The character embeddings of the word ``Mars'' are given to a bidirectional LSTMs. We concatenate their last outputs to an embedding from a lookup table to obtain a representation for this word.}
  \label{fig:char-model}
\end{figure}

An important distinction of our work from most previous approaches is that we learn character-level features while training instead of hand-engineering prefix and suffix information about words. Learning character-level embeddings has the advantage of learning representations specific to the task and domain at hand. They have been found useful for morphologically rich languages and to handle the out-of-vocabulary problem for tasks like part-of-speech tagging and language modeling \cite{ling:2015} or dependency parsing \cite{lstmemnlp15}.

Figure~\ref{fig:char-model} describes our architecture to generate a word embedding for a word from its characters. A character lookup table initialized at random contains an embedding for every character. The character embeddings corresponding to every character in a word are given in direct and reverse order to a forward and a backward LSTM. The embedding for a word derived from its characters is the concatenation of its forward and backward representations from the bidirectional LSTM. This character-level representation is then concatenated with a word-level representation from a word lookup-table. During testing, words that do not have an embedding in the lookup table are mapped to a UNK embedding. To train the UNK embedding, we replace singletons with the UNK embedding with a probability $0.5$. In all our experiments, the hidden dimension of the forward and backward character LSTMs are $25$ each, which results in our character-based representation of words being of dimension $50$.

Recurrent models like RNNs and LSTMs are capable of encoding very long sequences, however, they have a representation biased towards their most recent inputs. As a result, we expect the final representation of the forward LSTM to be an accurate representation of the suffix of the word, and the final state of the backward LSTM to be a better representation of its prefix. \ignore{Using a bidirectional LSTM to encode character representation of words, compared to a single LSTM, gives a small but significant improvement.}Alternative approaches---most notably like convolutional networks---have been proposed to learn representations of words from their characters \cite{zhang2015character,DBLP:journals/corr/KimJSR15}. However, convnets are designed to discover position-invariant features of their inputs. While this is appropriate for many problems, e.g., image recognition (a cat can appear anywhere in a picture), we argue that important information is position dependent (e.g., prefixes and suffixes encode different information than stems), making LSTMs an \emph{a priori} better function class for modeling the relationship between words and their characters.

\subsection{Pretrained embeddings}\label{sec:pretrained}
As in \newcite{collobert2011natural}, we use pretrained word embeddings to initialize our lookup table. We observe significant improvements using pretrained word embeddings over randomly initialized ones. Embeddings are pretrained using skip-n-gram \cite{skipngram}, a variation of word2vec \cite{mikolov2013efficient} that accounts for word order. These embeddings are fine-tuned during training.

Word embeddings for Spanish, Dutch, German and English are trained using the Spanish Gigaword version~3, the Leipzig corpora collection, the German monolingual training data from the 2010 Machine Translation Workshop and the English Gigaword version~4 (with the LA Times and NY Times portions removed) respectively.\footnote{\cite{graff2011spanish,biemann2007leipzig,callison2010findings,englishgigaword}} We use an embedding dimension of $100$ for English, $64$ for other languages, a minimum word frequency cutoff of $4$, and a window size of $8$.

\ignore{
Spanish:
          Lines: 34,145,390
         Tokens: 1,066,774,432
          Types: 3,257,431
     Singletons: 1,675,131

Dutch
          Lines: 1,000,006
         Tokens: 17,645,554
          Types: 650,980
     Singletons: 415,383

German
          Lines: 46,059,414
         Tokens: 817,561,850
          Types: 7,134,443
     Singletons: 4,110,035

English
          Lines: 24,517,385
         Tokens: 593,524,637
          Types: 1,588,637
     Singletons: 808,462
}

\subsection{Dropout training}\label{sec:dropout}
Initial experiments showed that character-level embeddings did not improve our overall performance when used in conjunction with pretrained word representations. To encourage the model to depend on both representations, we use dropout training \cite{dropout}, applying a dropout mask to the final embedding layer just before the input to the bidirectional LSTM in Figure \ref{fig:bilstm-crf}. We observe a significant improvement in our model's performance after using dropout (see table~\ref{results-diff-config}).

\section{Experiments}\label{sec:experiments}

This section presents the methods we use to train our models, the results we obtained on various tasks and the impact of our networks' configuration on model performance.

\subsection{Training}
For both models presented, we train our networks using the back-propagation algorithm updating our parameters on every training example, one at a time, using stochastic gradient descent (SGD) with a learning rate of $0.01$ and a gradient clipping of $5.0$. Several methods have been proposed to enhance the performance of SGD, such as Adadelta \cite{adadelta} or Adam \cite{adam}. Although we observe faster convergence using these methods, none of them perform as well as SGD with gradient clipping.
\guillaumecomment{the 2 sentences above are maybe not super important}

Our LSTM-CRF model uses a single layer for the forward and backward LSTMs whose dimensions are set to $100$. Tuning this dimension did not significantly impact model performance. We set the dropout rate to $0.5$. Using higher rates negatively impacted our results, while smaller rates led to longer training time.

The stack-LSTM model uses two layers each of dimension $100$ for each stack. The embeddings of the actions used in the composition functions
have $16$ dimensions each, and the output embedding is of dimension $20$. We experimented with different dropout rates and reported the scores using the best dropout rate for each language.\footnote{English (D=$0.2$), German, Spanish and Dutch (D=$0.3$)} It is a greedy model that apply locally optimal actions until the entire sentence is processed, further improvements might be obtained with beam search \cite{zhang:2011} or training with exploration \cite{ballesteros-arxiv16}.

\subsection{Data Sets}

We test our model on different datasets for named entity recognition. To demonstrate our model's ability to generalize to different languages, we present results on the CoNLL-2002 and CoNLL-2003 datasets \cite{TjongKimSang:2002:ICS:1118853.1118877,TjongKimSang-DeMeulder:2003:CONLL} that contain independent named entity labels for English, Spanish, German and Dutch. All datasets contain four different types of named entities: locations, persons, organizations, and miscellaneous entities that do not belong in any of the three previous categories. \ignore{We used the CoNLL-2000 dataset to evaluate our models for chunking. }Although POS tags were made available for all datasets, we did not include them in our models. We did not perform any dataset preprocessing, apart from replacing every digit with a zero in the English NER dataset.


\subsection{Results}
Table~\ref{results-ner-en} presents our comparisons with other models for named entity recognition in English. To make the comparison between our model and others fair, we report the scores of other models with and without the use of external labeled data such as gazetteers and knowledge bases. Our models do not use gazetteers or any external labeled resources. The best score reported on this task is by \newcite{luojoint}. They obtained a F$_1$ of 91.2\ignore{let's decide that later\guillaumecomment{I disagree, maybe they just have 91.15} \miguelcomment{is there any difference?}} by jointly modeling the NER and entity linking tasks \cite{hoffart2011robust}. Their model uses a lot of hand-engineered features including spelling features, WordNet clusters, Brown clusters, POS tags, chunks tags, as well as stemming and external knowledge bases like Freebase and Wikipedia. Our LSTM-CRF model outperforms all other systems, including the ones using external labeled data like gazetteers. Our Stack-LSTM model also outperforms all previous models that do not incorporate external features, apart from the one presented by \newcite{chiu2015named}.

Tables~\ref{results-ner-de},~\ref{results-ner-nld} and~\ref{results-ner-es} present our results on NER for German, Dutch and Spanish respectively in comparison to other models. On these three languages, the LSTM-CRF model significantly outperforms all previous methods, including the ones using external labeled data. The only exception is Dutch, where the model of \newcite{gillick2015multilingual} can perform better by leveraging the information from other NER datasets. The Stack-LSTM also consistently presents state-the-art (or close to) results compared to systems that do not use external data.

As we can see in the tables, the Stack-LSTM model is more dependent on character-based representations to achieve competitive performance; we hypothesize that the LSTM-CRF model requires less orthographic information since it gets more contextual information out of the bidirectional LSTMs; however, the Stack-LSTM model consumes the words one by one and it just relies on the word representations when it chunks words.

\ignore{
Table~\ref{results-chunk} compares different results on chunking. The models used by \cite{collobert2011natural} and \cite{huang:2015} are the same as the ones they used for English NER. \cite{shen2005voting} uses a majority vote on the output of different specialized HMMs trained on different tagging schemes (IOB1, IOB2, etc.).
}

\begin{table}[!ht]
\centering
\begin{scriptsize}
\begin{tabular}{l|c}
\textbf{Model} & \textbf{F}${_{\mathbf{1}}}$ \\
\hline
\newcite{collobert2011natural}* & 89.59 \\
\newcite{lin2009phrase} & 83.78 \\
\newcite{lin2009phrase}* & 90.90 \\
\newcite{huang:2015}* & 90.10 \\
\newcite{passos2014lexicon} & 90.05 \\
\newcite{passos2014lexicon}* & 90.90 \\
\newcite{luojoint}* + gaz & 89.9 \\
\newcite{luojoint}* + gaz + linking & \bf91.2 \\
\newcite{chiu2015named} & 90.69 \\
\newcite{chiu2015named}* & 90.77 \\
\hline
\hline
LSTM-CRF (no char) & 90.20\\
LSTM-CRF & \textbf{90.94}\\
S-LSTM (no char) & 87.96\\
S-LSTM & 90.33\\
\end{tabular}
\end{scriptsize}
\caption{English NER results (CoNLL-2003 test set). *~indicates models trained with the use of external labeled data\guillaumecomment{I'm not sure the 89.9 Luo et al. report is without external data, it's unclear in the paper. Also I do not like the table, it's not super nice. I do not see how to make it nicer} \miguelcomment{I think they clearly use external data. If you read their section 4.3, they say that the 89.9 result is obtained only with NER features which are described in 4.3.1. In 4.3.1 they say that they use among other things... We collect several
known name lists, like popular English first/last names for people, organization lists and so on from Wikipedia and Freebase.}}
\label{results-ner-en}
\end{table}%

\begin{table}[!ht]
\centering
\begin{scriptsize}
\begin{tabular}{l|c}
\textbf{Model} & \textbf{F}${_{\mathbf{1}}}$ \\
\hline
\newcite{florian2003named}* & 72.41 \\
\newcite{ando2005framework} & 75.27 \\
\newcite{qi2009combining} & 75.72 \\
\newcite{gillick2015multilingual} & 72.08 \\
\newcite{gillick2015multilingual}* & 76.22 \\
\hline
\hline
LSTM-CRF -- no char & 75.06 \\
LSTM-CRF & \bf78.76 \\
S-LSTM -- no char & 65.87 \\
S-LSTM & 75.66 \\
\end{tabular}
\end{scriptsize}
\caption{German NER results (CoNLL-2003 test set). *~indicates models trained with the use of external labeled data\guillaumecomment{Score on stack lstm without char seems anormally too low here} \miguelcomment{I tried several versions of it... and it does not get better...I'm still training a model and will ping you if I find something better.}}
\label{results-ner-de}
\end{table}%

\begin{table}[!ht]
\centering
\begin{scriptsize}
\begin{tabular}{l|c}
\textbf{Model} & \textbf{F}${_{\mathbf{1}}}$ \\
\hline
\newcite{carreras2002named} & 77.05 \\
\newcite{nothman2013learning} & 78.6 \\
\newcite{gillick2015multilingual} & 78.08 \\
\newcite{gillick2015multilingual}* & \bf82.84 \\
\hline
\hline
LSTM-CRF -- no char & 73.14 \\
LSTM-CRF & \bf81.74 \\
S-LSTM -- no char & 69.90 \\
S-LSTM & 79.88 \\
\end{tabular}
\end{scriptsize}
\caption{Dutch NER (CoNLL-2002 test set). *~indicates models trained with the use of external labeled data}
\label{results-ner-nld}
\end{table}%

\begin{table}[!ht]
\centering
\begin{scriptsize}
\begin{tabular}{l|c}
\textbf{Model} & \textbf{F}${_{\mathbf{1}}}$ \\
\hline
\newcite{carreras2002named}* & 81.39 \\
\newcite{santos2015boosting} & 82.21 \\
\newcite{gillick2015multilingual} & 81.83 \\
\newcite{gillick2015multilingual}* & 82.95 \\
\hline
\hline
LSTM-CRF -- no char & 83.44 \\
LSTM-CRF & \bf85.75 \\
S-LSTM -- no char & 79.46\\
S-LSTM & 83.93\\
\end{tabular}
\end{scriptsize}
\caption{Spanish NER (CoNLL-2002 test set). *~indicates models trained with the use of external labeled data}
\label{results-ner-es}
\end{table}%

\ignore{
\begin{table}[!ht]
\centering
\begin{tabular}{l|c}
\textbf{Model} & \textbf{F}${_{\mathbf{1}}}$ \\
\hline
\newcite{shen2005voting} & 94.01 \\
\newcite{collobert2011natural} & 94.32 \\
\newcite{huang:2015} & 94.46 \\
\hline
\hline
LSTM-CRF - no char & 94.62 \\
LSTM-CRF & 94.62 \\
S-LSTM & XX \\
\end{tabular}
\caption{Chunking (CoNLL-2000)}
\label{results-chunk}
\end{table}%
}

\subsection{Network architectures}

Our models had several components that we could tweak to understand their impact on the overall performance. We explored the impact that the CRF, the character-level representations, pretraining of our word embeddings and dropout had on our LSTM-CRF model. We observed that pretraining our word embeddings gave us the biggest improvement in overall performance of $+7.31$ in F$_1$. The CRF layer gave us an increase of $+1.79$, while using dropout resulted in a difference of $+1.17$ and finally learning character-level word embeddings resulted in an increase of about $+0.74$. For the Stack-LSTM we performed a similar set of experiments. Results with different architectures are given in table~\ref{results-diff-config}.

\begin{table}[h]
\centering
\begin{scriptsize}
\begin{tabular}{l|l|c}
\textbf{Model} & \textbf{Variant} & \textbf{F}${_{\mathbf{1}}}$\\
\hline
LSTM & char + dropout + pretrain & 89.15 \\
LSTM-CRF & char + dropout & 83.63 \\
LSTM-CRF & pretrain & 88.39 \\
LSTM-CRF & pretrain + char & 89.77 \\
LSTM-CRF & pretrain + dropout & 90.20 \\
LSTM-CRF & pretrain + dropout + char & \bf90.94 \\
\hline
S-LSTM & char + dropout & 80.88 \\
S-LSTM & pretrain & 86.67 \\
S-LSTM & pretrain + char & 89.32 \\
S-LSTM & pretrain + dropout & 87.96 \\
S-LSTM & pretrain + dropout + char & 90.33 \\
\end{tabular}
\end{scriptsize}
\caption{English NER results with our models, using different configurations. ``pretrain'' refers to models that include pretrained word embeddings, ``char'' refers to models that include character-based modeling of words, ``dropout'' refers to models that include dropout rate.}
\label{results-diff-config}
\end{table}%

\section{Related Work}
\label{relwork}


In the CoNLL-2002 shared task, \newcite{carreras2002named} obtained among the best results on both Dutch and Spanish by combining several small fixed-depth decision trees. Next year, in the CoNLL-2003 Shared Task, \newcite{florian2003named} obtained the best score on German by combining the output of four diverse classifiers. \newcite{qi2009combining} later improved on this with a neural network by doing unsupervised learning on a massive unlabeled corpus.\ignore{ Both of these models use spelling features, and \newcite{florian2003named} used an external list of gazetteers.}

Several other neural architectures have previously been proposed for NER. For instance, \newcite{collobert2011natural} uses a CNN over a sequence of word embeddings with a CRF layer on top. This can be thought of as our first model without character-level embeddings and with the bidirectional LSTM being replaced by a CNN. More recently, \newcite{huang:2015} presented a model similar to our LSTM-CRF, but using hand-crafted spelling features. \newcite{zhou2015end} also used a similar model and adapted it to the semantic role labeling task. \newcite{lin2009phrase} used a linear chain CRF with $L_2$ regularization, they added phrase cluster features extracted from the web data and spelling features. \newcite{passos2014lexicon} also used a linear chain CRF with spelling features and gazetteers.

Language independent NER models like ours have also been proposed in the past. Cucerzan and Yarowsky \shortcite{cucerzan1999language,cucerzan2002language} present semi-supervised bootstrapping algorithms for named entity recognition by co-training character-level (word-internal) and token-level (context) features. \newcite{eisenstein2011structured} use Bayesian nonparametrics to construct a database of named entities in an almost unsupervised setting. \newcite{ratinov2009design} quantitatively compare several approaches for NER and build their own supervised model using a regularized average perceptron and aggregating context information.

\ignore{\newcite{hammerton2003named} present recurrent neural architectures for NER without learning character-level representations.}

Finally, there is currently a lot of interest in models for NER that use letter-based representations. \newcite{gillick2015multilingual} model the task of sequence-labeling as a sequence to sequence learning problem and incorporate character-based representations into their encoder model. \newcite{chiu2015named} employ an architecture similar to ours, but instead use CNNs to learn character-level features, in a way similar to the work by \newcite{santos2015boosting}. 


\section{Conclusion}
This paper presents two neural architectures for sequence labeling that provide the best NER results ever reported in standard evaluation settings, even compared with models that use external resources, such as gazetteers.

A key aspect of our models are that they model output label dependencies, either via a simple CRF architecture, or using a transition-based algorithm to explicitly construct and label chunks of the input. Word representations are also crucially important for success: we use both pre-trained word representations and ``character-based'' representations that capture morphological and orthographic information. To prevent the learner from depending too heavily on one representation class, dropout is used.

\section*{Acknowledgments}
This work was sponsored in part by the Defense Advanced Research Projects Agency (DARPA)
Information Innovation Office (I2O) under the Low Resource Languages for Emergent Incidents (LORELEI) program issued by DARPA/I2O under Contract No.~HR0011-15-C-0114. Miguel Ballesteros is supported by the European Commission under the contract numbers FP7-ICT-610411 (project MULTISENSOR) and H2020-RIA-645012 (project KRISTINA).

\bibliographystyle{naaclhlt2016}
\bibliography{biblio}

\end{document}